\documentclass{ceurart}
\usepackage{tikz}
\usetikzlibrary{positioning,arrows.meta,shapes.geometric,fit,backgrounds,calc}

\usepackage[most]{tcolorbox}
\tcbuselibrary{skins,breakable}

\newtcolorbox{promptbox}[1]{%
  enhanced,
  breakable,
  colback=white,
  colframe=black,
  boxrule=0.4pt,
  arc=2pt,
  left=6pt, right=6pt, top=6pt, bottom=6pt,
  fonttitle=\bfseries\sffamily\small,
  coltitle=white,
  colbacktitle=black,
  attach boxed title to top left={xshift=0pt, yshift=0pt},
  boxed title style={
    colback=black,
    colframe=black,
    boxrule=0pt,
    arc=2pt,
    left=6pt, right=6pt, top=2pt, bottom=2pt,
  },
  title={#1},
}

\copyrightyear{2026}
\copyrightclause{Copyright for this paper by its authors.
  Use permitted under Creative Commons License Attribution 4.0
  International (CC BY 4.0).}

\conference{CLEF 2026: Conference and Labs of the Evaluation Forum,
  September 21--24, 2026, Jena, Germany}

\begin{document}

\title{UZHCL\_for\_the\_win at Touch\'{e}: Encoder Fine-Tuning, Synthetic Augmentation, and an Underpowered Encoder--LLM Comparison}
\title[mode=sub]{Touch\'{e} at CLEF 2026}

\author[1]{Stylianos Psychias}[%
  email=stylianos.psychias@uzh.ch,
]
\address[1]{University of Zurich, Zurich, Switzerland}

\begin{abstract}
Online discussions on platforms like Reddit are a productive testbed for fallacy detection: short texts, broad topical range, and the full inventory of informal fallacies catalogued by classical argumentation theory long before social media. The Touch\'{e} 2026 lab at CLEF formalises this as a shared task on a Reddit-derived corpus and provides a structured benchmark for systems that combine fallacy detection with argument-scheme classification. This paper describes our participation in the lab, covering all three subtasks (ST1 binary fallacy detection, ST2 eight-way fallacy classification, ST3 four-way argument-scheme classification) and both evaluation tracks. Our submitted systems are fine-tuned RoBERTa-large encoders with task-specific heads, augmented with a small batch of LLM-generated synthetic contrastive pairs; for ST2 and ST3 on the enhanced track we additionally extended training with pseudo-labelled examples, extra training items that the encoder labels itself from the published unlabelled pool, kept only where it is most confident and only after a verification check has passed. As an alternative we ran a wider backbone comparison spanning additional encoders (DeBERTa-v3, ModernBERT) and an LLM track (four Qwen variants from 3B to 32B, zero-shot through CoT-LoRA); at this corpus scale none showed a validated benefit over the RoBERTa-large encoder, with 5-fold variance wide enough that we call the comparison underpowered rather than settled. The enhanced track exposes organiser-rewritten input fields (scored separately from the base track); we characterise how much of its lift reflects the rewrite carrying label information rather than genuine argument clarification, and anchor cross-comparisons on base-track or 5-fold CV behaviour. On our held-out development partitions the submitted systems perform strongly on the enhanced and binary cells, with the two base-track classification cells underestimated by their small dev splits. On the official Touch\'{e} 2026 test set they ranked first on both base-track classification cells (ST2 and ST3) and second on both binary-detection (ST1) tracks, with the two enhanced-track classification cells topped by cascaded and multitask systems. We close on the practical-external rare-class limitation on ST3, the limitations of the present work, and the most useful follow-ups. All code is publicly available at \url{https://github.com/psychias/fallacy_detection_clef_2026}.
\end{abstract}


\maketitle

\section{Introduction}
\label{sec:intro}
Online discussions on platforms like Reddit are a productive testbed for fallacy detection: the texts are short, the topics range across politics, science, and lifestyle, and the rhetorical moves include the full inventory of informal fallacies that classical argumentation theory catalogued long before social media~\cite{sahai2021invisiblewall,walton2008schemes}. The Touch\'{e} 2026 lab at CLEF formalises this as a shared task on a Reddit-derived corpus and provides a structured benchmark for systems that combine fallacy detection with argument-scheme classification~\cite{kiesel:2026c}.

Our submission and this notebook contribute to the lab's overall record as documented in the Touch\'{e} 2026 overview paper~\cite{kiesel:2026c} and its extended companion~\cite{kiesel:2026d}.

We participated in all three subtasks of the 2026 edition~\cite{kiesel:2026c}: binary fallacy detection (ST1), eight-way fallacy classification (ST2), and four-way argument-scheme classification (ST3) organised along a goal axis (practical vs.\ epistemic) crossed with a basis axis (internal vs.\ external)~\cite{macagno2022argumentation}. Each subtask is evaluated under two tracks. The \emph{base} track uses only original Reddit fields. The \emph{enhanced} track additionally exposes organiser-supplied paraphrases that were deliberately rewritten to surface fallacy and scheme structure as a designed difficulty setting; the two tracks are scored separately, so the enhanced rewrite is a sanctioned aid, not a contamination of base-track scoring. We submit one run per subtask and track, totalling six TIRA runs~\cite{froebe:2023}.

\paragraph{Contributions.} The paper makes four contributions, which the methodology and results sections map onto one-to-one:

\begin{itemize}
  \item[\textbf{C1}] A uniform RoBERTa-large~\cite{liu2019roberta} fine-tuning pipeline applied across all six (subtask, track) cells with minimal per-cell variation (\S\ref{sec:pipeline}, \S\ref{sec:backbone}).
  \item[\textbf{C2}] An augmentation study combining LLM-generated synthetic contrastive pairs with a gated pseudo-labelling top-up of the published unlabelled pool, with the negative results (a pseudo-label loss on ST1 enhanced, a rare-class top-up that hurt ST3 base) reported alongside the gains (\S\ref{sec:synth}, \S\ref{sec:pseudo}).
  \item[\textbf{C3}] A backbone comparison covering DeBERTa-v3-base/large~\cite{he2021debertav3}, ModernBERT-large~\cite{warner2024modernbert}, and four Qwen variants from 3B to 32B parameters~\cite{qwen25,qwen3} fine-tuned with LoRA / QLoRA~\cite{hu2022lora,dettmers2023qlora} and chain-of-thought supervision~\cite{wei2022cot}, showing no validated benefit over the encoder at this corpus scale (\S\ref{sec:llm-detour}, \S\ref{sec:disc-encoder}).
  \item[\textbf{C4}] A characterisation of two structural features of the 2026 evaluation: what the enhanced-track rewrite actually contributes to a classifier (a string-mask sanity check, a per-class distribution that is consistent with strong label-bearing signal, and a falsifiable follow-up; \S\ref{sec:disc-leakage}), and the practical-external (\textsf{pe}) rare-class limitation that survived targeted augmentation, threshold calibration, and a basis-axis meta-learner (\S\ref{sec:disc-pe}).
\end{itemize}

A central posture across all four is careful reading of small-corpus comparisons, separating what dev numbers and 5-fold CV evidence can support from what they cannot. We keep the negative results in the body rather than in footnotes, because the design space they map out is the most useful contribution this notebook can offer a team considering the same shared task next year.

The rest of the paper is organised as follows. Section~\ref{sec:background} situates the work inside the fallacy-detection and argument-scheme NLP lineage. Section~\ref{sec:system} describes the pipeline, the backbone selection, and the LLM alternative we evaluated. Section~\ref{sec:results} reports per-subtask numbers, the augmentation ablation, and the backbone comparison. Section~\ref{sec:discussion} discusses the encoder--LLM comparison, the enhanced-track signal, the \textsf{pe} limitation, the limitations of the present work, and the most useful follow-ups. Section~\ref{sec:conclusion} summarises our conclusions.

\section{Previous Work}
\label{sec:background}
\subsection{Fallacy detection in NLP}
\label{sec:bg-fallacy}

The Reddit informal-fallacy dataset that the Touch\'{e} 2026 lab redistributes was introduced by Sahai et al.~\cite{sahai2021invisiblewall}, who labelled a long-tail eight-way taxonomy over comments mined from r/changemyview and adjacent subreddits and reported the first transformer baselines on the resulting corpus. Adjacent benchmarks treat closely related phenomena with similar fine-grained label structures. The Argument Reasoning Comprehension task introduced multi-class warrant identification on user-written argument pairs and surfaced how brittle large language models can be on argumentation tasks that require implicit-premise reasoning~\cite{habernal2018arc}. Da San Martino et al.~\cite{dasanmartino2019propaganda} catalogued eighteen propaganda techniques in news articles. Several of those techniques (\emph{appeal to authority}, \emph{appeal to fear}, \emph{black-and-white fallacy}) overlap directly with the ST2 inventory and indicate that the propaganda-detection literature is a useful cross-domain reference even though the source genre differs. Goffredo et al.~\cite{goffredo2022political} classified fallacies in political-debate transcripts; their taxonomy differs from the ST2 inventory, but their supervised setting is a close methodological neighbour for the classifier we build.

Two more recent papers anchor the picture. Alhindi et al.~\cite{alhindi2022multitask} reported strong multitask instruction-prompted results across a suite of fallacy-recognition datasets, demonstrating that multitask supervision over related fallacy datasets transfers usefully to eight-way tasks of this kind. MAFALDA~\cite{helwe2024mafalda} is the most-cited 2024 unified fallacy benchmark, we do not retrain on MAFALDA, but our ST2 label inventory is recognisable in its taxonomy and the MAFALDA error-analysis findings on rare-class behaviour echo what we report for the ST3 \textsf{pe} class in \S\ref{sec:disc-pe}. Pan et al.~\cite{pan2024zeroshotfallacy} provide the published precedent for evaluating LLMs as zero-shot fallacy classifiers. They show that zero-shot LLM performance is real but uneven across fallacy types, which motivates the LLM track we describe in \S\ref{sec:llm-detour} as worth running rather than dismissing.

\subsection{Argumentation schemes}
\label{sec:bg-schemes}

The four ST3 labels are not ad hoc, but neither do they map one-to-one onto individual Walton schemes. The goal (practical / epistemic) $\times$ basis (internal / external) crossing is the lab's operationalisation~\cite{kiesel:2026c} of the argumentation-profile framing of Macagno~\cite{macagno2022argumentation}, itself rooted in the broader Walton--Reed--Macagno scheme tradition~\cite{walton2008schemes}. The crossing summarises, at a coarser grain than a full scheme catalogue, whether the speaker is arguing about what to do or what to believe, and whether the warrant comes from the speaker's own resources or from an external source. NLP work that classifies arguments by Walton scheme is comparatively thin: Feng and Hirst~\cite{feng2011scheme} gave the canonical treatment, assigning arguments to five common schemes, while Jo et al.~\cite{jo2021schemes} classify argumentative \emph{relations} using scheme-based logical mechanisms rather than scheme labels. Neither targets the ST3 goal/basis label set directly, so a system designer cannot transfer a published scheme-classification recipe onto this task.

The Touch\'{e} lab itself has progressively shifted over recent editions from argument retrieval~\cite{bondarenko2022touche,bondarenko2023touche} to full argumentation systems~\cite{kiesel2024touche,kiesel2025touche}, with the 2026 edition the first to feature fallacy detection as a dedicated task. The encoder-vs-LLM trade-off for small-corpus fallacy classification is not, to our reading, settled by the published literature: encoder fine-tuning, LoRA-adapted mid-size LLMs, and zero/few-shot prompted frontier models have each been reported as competitive on neighbouring tasks, with no clean recipe-matched comparison at the corpus scale of the present shared task. That open question is the gap our contribution C3 addresses, and it motivates the systematic backbone comparison reported in \S\ref{sec:results}.

\section{Methodology}
\label{sec:system}
\subsection{Pipeline overview}
\label{sec:pipeline}

The pipeline has three input streams feeding one fine-tuned encoder: real examples, synthetic contrastive pairs, and a gated pseudo-labelled top-up --- a small batch of extra training items that the encoder labels itself from the published unlabelled data, kept only for the examples it is most confident about and only after a verification check has passed. \S\ref{sec:backbone} fixes the backbone (supporting C1), \S\ref{sec:synth}--\S\ref{sec:pseudo} build the two augmented streams (supporting C2), and \S\ref{sec:llm-detour} reports the LLM alternative we evaluated against this pipeline (supporting C3). Our main focus this year was on building a uniform fine-tuning pipeline that we could apply, with minimal cell-specific variation, across every (subtask, track) combination the lab requires. The Touch\'{e} 2026 lab~\cite{kiesel:2026c} (task description and labels at \url{https://touche.webis.de/clef26/touche26-web/fallacy-detection.html}) defines three subtasks over the Reddit informal-fallacy data of Sahai et al.~\cite{sahai2021invisiblewall}, with evaluation hosted on TIRA~\cite{froebe:2023}. Subtask ST1 is a binary decision on whether a given comment, presented together with its parent comment and the title of the originating thread, commits any informal fallacy. Subtask ST2 conditions on a positive ST1 decision and asks for the specific fallacy type from an eight-way inventory (using the official label strings from the task schema): \textsf{authority}, \textsf{black-white}, \textsf{hasty-generalization}, \textsf{natural}, \textsf{population}, \textsf{slippery-slope}, \textsf{tradition}, \textsf{worse-problems}. Subtask ST3 conditions on a negative ST1 decision and asks for the argument scheme of the comment from a four-way inventory~\cite{walton2008schemes,macagno2022argumentation}: \textsf{practical-internal}, \textsf{practical-external}, \textsf{epistemic-internal}, \textsf{epistemic-external}, which we abbreviate \textsf{pi}, \textsf{pe}, \textsf{ei}, \textsf{ee} respectively in the rest of the paper. Evaluation is by macro-averaged F1 in all three subtasks, which puts roughly equal weight on each label and turns rare-class behaviour into a binding constraint on the headline number, a point we return to in \S\ref{sec:disc-pe}. Each subtask is evaluated in two tracks. The \emph{base} track exposes only original Reddit fields (in our setup, \texttt{text\_raw}, \texttt{text\_raw\_parent}, \texttt{text\_raw\_title}, \texttt{text\_base}, \texttt{argument\_base}). The \emph{enhanced} track additionally exposes two organiser-supplied paraphrases, \texttt{text\_enhanced} and \texttt{argument\_enhanced}, that were rewritten with knowledge of the gold labels in order to make the argumentative structure of the comment more explicit, and is scored as a separate track from the base submissions. The rewrite is a designed feature; we defer the characterisation of how much label-bearing signal it carries to \S\ref{sec:disc-leakage}, where the analysis sits in one place. The lab permits up to three runs per (subtask, track), for a maximum of nine submissions. We used six, populating one slot per subtask per track. Appendix~\ref{app:citgraph} traces which prior work motivated each component of the pipeline described below.

\begin{figure}[h]
\centering
\begin{tikzpicture}[
  every node/.style={font=\scriptsize, align=center},
  box/.style={rectangle, draw=black!70, line width=0.4pt,
              rounded corners=2pt, inner sep=2pt,
              minimum width=20mm, minimum height=8mm},
  inp/.style={box, fill=black!4},
  proc/.style={box, fill=black!8},
  arr/.style={-{Latex[length=1.6mm]}, line width=0.4pt},
  lbl/.style={midway, sloped, above, font=\tiny, inner sep=1pt}
]
\node[inp] (real)    at (0,  1.8) {Real\\examples};
\node[inp] (synth)   at (0,  0)   {Synthetic\\contrastive pairs};
\node[inp] (pool)    at (0, -1.8) {Unlabelled pool\\(enhanced only)};
\node[proc] (gate)   at (3.0, -1.8) {Audit gate\\\texttt{exp\_0081}};
\node[inp]  (pseudo) at (6.0, -1.8) {Pseudo-\\labelled};
\node[proc] (mix)    at (6.0,  0)   {Mini-batch\\mix};
\node[proc] (enc)    at (9.0,  0)   {RoBERTa-large\\encoder};
\node[proc] (head)   at (12.0, 0)   {Per-task head\\(argmax)};
\draw[arr] (real)   -- node[lbl] {\texttt{sw=1.0}} (mix);
\draw[arr] (synth)  -- node[midway, above, font=\tiny, inner sep=1pt] {\texttt{sw=0.5}} (mix);
\draw[arr] (pool)   -- (gate);
\draw[arr] (gate)   -- (pseudo);
\draw[arr] (pseudo) -- node[lbl] {per-cell \texttt{sw}} (mix);
\draw[arr] (mix)  -- (enc);
\draw[arr] (enc)  -- (head);
\end{tikzpicture}
\caption{Training and inference pipeline for the submitted RoBERTa-large systems.
Real, synthetic, and pseudo-labelled examples are mixed at per-stream loss weights
(real at 1.0; the synthetic and pseudo-labelled streams down-weighted per cell,
see Appendix~\ref{app:hyper}) into a single fine-tuning loop. The audit gate sits between
pseudo-label generation and pseudo-label training (\S\ref{sec:pseudo}). At inference time the trained
classifier head is decoded with per-class thresholds for ST3 and default argmax
for ST1/ST2.}
\label{fig:pipeline}
\end{figure}
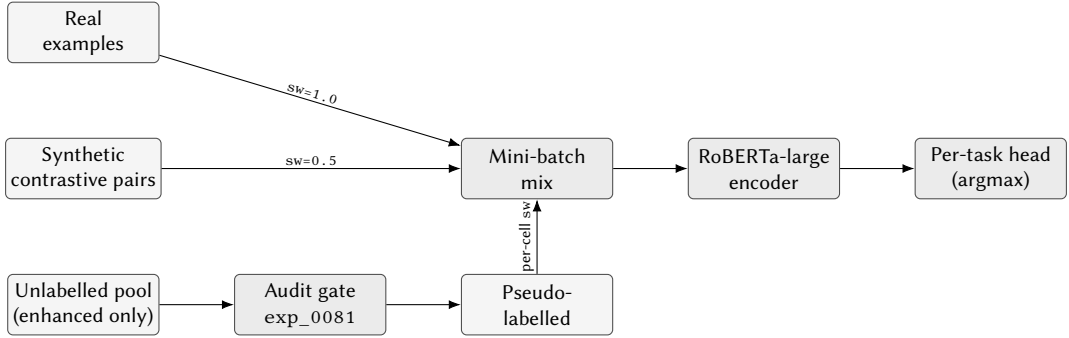

The same pipeline produces all six submitted runs, with per-cell hyperparameters summarised in Appendix~\ref{app:hyper}. Real examples are concatenated from the published Reddit fields, augmented with a small batch of LLM-generated synthetic contrastive pairs (\S\ref{sec:synth}), and for ST2 and ST3 enhanced extended with pseudo-labelled unlabelled examples (\S\ref{sec:pseudo}) once the dev-overlap audit has passed. The three input streams are interleaved in each training mini-batch under the per-stream loss weights given in the caption. The mixed corpus is fed into a RoBERTa-large encoder~\cite{liu2019roberta} with a task-specific classification head. Loss is class-weighted cross-entropy for ST2 and ST3 (with class weights set to inverse-frequency on the training partition) and unweighted cross-entropy for ST1. We use the HuggingFace \texttt{transformers} library throughout~\cite{wolf2020transformers}, with the AdamW optimiser, a linear warmup over the first 10\% of steps, and a peak learning rate of 1e-5. Inference uses default argmax over class logits for all three subtasks, including the submitted ST3 runs. We additionally explored a per-class threshold sweep on the ST3 development partition (\S\ref{sec:thresholds}) as a diagnostic, but did not apply it to the uploaded predictions. The same training loop is used for all six (subtask, track) cells. Only the input fields, the loss weights, the truncation length, and the presence or absence of pseudo-labelled examples vary across cells. Inputs are concatenated using the standard RoBERTa special tokens \texttt{<s>...</s></s>...</s>}, with tokeniser truncation set to \texttt{max\_len=384} for ST3, \texttt{max\_len=256} for ST1 enhanced and both ST2 cells, and \texttt{max\_len=512} for the submitted ST1-base run (\texttt{exp\_0161}), which concatenates \texttt{text\_raw} with \texttt{text\_base}. The choice between these windows was driven by a token-length diagnostic showing that ST3 inputs benefited measurably from the longer window once \texttt{argument\_enhanced} was concatenated, while the shorter ST1-enhanced and ST2 inputs fit comfortably inside 256 with only a small tail of truncations. 

\subsection{Backbone selection}
\label{sec:backbone}

We compared RoBERTa-base, RoBERTa-large, DeBERTa-v3-base, DeBERTa-v3-large, and ModernBERT-large under matched data and training budgets. RoBERTa-large beat RoBERTa-base by 0.027 on ST1 base, 0.007 on ST1 enhanced, and 0.055 on ST2 base under matched recipes. 
The one cell where RoBERTa-base was marginally ahead, ST2 enhanced (0.925 vs.\ 0.923), sits inside seed-variance for these models and we treat it as a tie rather than evidence against the larger backbone. 
We therefore standardised on RoBERTa-large for the six submitted runs.

DeBERTa-v3-large did not survive the fine-tuning recipes we tried. Plain dev runs collapsed to near-chance F1 across all three subtasks (ST1 base 0.351, ST2 base 0.030, ST3 base 0.205 across two attempts). 
Mixed-precision training NaN-exploded across four subtask$\times$sw combinations. Switching to bf16 with a smaller learning rate stopped the NaNs but produced near-zero F1. 
After four NaN-explosion incidents and seven collapse-or-fail runs, we removed DeBERTa-v3-large from the backbone shortlist.

ModernBERT-large~\cite{warner2024modernbert} ran cleanly but did not beat RoBERTa-large on ST3 enhanced, reaching 0.719 against RoBERTa-large's 0.735. 
A 0.016 F1 gap on a 4-way task is real but not enough to override the simpler, well-understood RoBERTa pipeline for a notebook-paper submission. For a corpus of roughly 473 ST3 examples, RoBERTa-large is the rational choice. We revisit the LLM alternatives in \S\ref{sec:llm-detour} and \S\ref{sec:disc-encoder}.

\subsection{Synthetic data augmentation}
\label{sec:synth}

We generated synthetic contrastive pairs using a frontier LLM via API. The lab data additionally ships a \texttt{resembles\_fallacy} field that pairs each non-fallacious comment with a structurally similar fallacy class, exactly the near-miss contrast our synthetic pairs are designed to manufacture; we did not consume that field as an explicit training signal, and treating it as a built-in contrastive supervision channel for ST1 is a natural follow-up we leave for a future submission~\cite{kiesel:2026c}. For each target label, we prompted the model with three real in-domain examples and asked it to produce a paired contrast, a minimally edited rewrite that flips the label while preserving topic, register, and Reddit-conversational style. Generations were filtered by a length sanity check and a string-overlap heuristic against the source pair to discourage trivial copies. Three batches were produced: an initial \texttt{synth\_v002} (130 pairs for ST1/ST2 and 110 pairs for ST3); a rare-class top-up \texttt{synth\_v003} for ST3 (100 epistemic-external + 50 practical-external pairs); and a separate \textsf{pe}-targeted batch (100 practical-external + practical-anchor pairs).\footnote{The exact prompt template and the generated pairs are released alongside this notebook in the TIRA submission package.} 

Synth at \texttt{sw=0.5} added 0.014--0.019 F1 across the four ST1 and ST2 cells (Table~\ref{tab:ablation}). 
On ST3 enhanced, however, synth alone did not move the needle: the real-only baseline and the +synth configuration tied at 0.694, and the eventual gain on this cell only appeared once pseudo-labelling was added (\S\ref{sec:pseudo}). 

The largest single gain came on ST3 base, where the \texttt{synth\_v002} recipe at \texttt{max\_len=384} lifted F1 from 0.436 to 0.523, an improvement of +0.087. 
Synth is not a free lunch: adding the \texttt{synth\_v003} rare-class top-up on top of \texttt{synth\_v002} dropped ST3-base F1 by 0.045 relative to \texttt{synth\_v002} alone, and a focal-loss recipe~\cite{lin2017focal} lost 0.018 over the same baseline. 
We read this as confirmation of the general caveat documented by Li et al.~\cite{li2023synthdata}: synthetic data improves the average and harms the worst-case, and which side you land on depends on the recipe.

\subsection{Pseudo-labelling}
\label{sec:pseudo}

For ST2 and ST3 enhanced we additionally added pseudo-labelled examples drawn from the published unlabelled pool. Predictions from the best dev-F1 checkpoint at the matching subtask $\times$ track were filtered by confidence at a nominal threshold $\tau=0.9$ (relaxed for under-populated classes so that rare schemes were not dropped entirely) and added to the training mix as an additional down-weighted stream; the per-cell loss weights are listed in Appendix~\ref{app:hyper}.\footnote{The pseudo-label confidence threshold $\tau$ and the source checkpoint are released in the TIRA reproducibility package alongside the prompt template.} Following the modern transformer-era treatment by Mukherjee and Awadallah~\cite{mukherjee2020ust} and the original pseudo-label idea of Lee~\cite{lee2013pseudolabel}, we treated this step with caution: pseudo examples are drawn from the unlabelled pool and could in principle be near-duplicates of dev instances, inflating the dev score, so we ran a dedicated dev-overlap audit (\texttt{exp\_0081}) before letting them into training. It checked id-level, exact-text, and token-Jaccard overlap between the pseudo-labelled entries and the dev split, and found no id or exact-text overlap and no token-Jaccard pair above 0.5 (maximum 0.30); we proceeded. 
The recipe paid off most clearly on ST2 enhanced, where adding pseudo lifted F1 from 0.937 to 0.970, a gain of +0.033 over the synth-only baseline. 
On ST3 enhanced the same recipe at \texttt{max\_len=384} lifted F1 from 0.694 to 0.735, a gain of +0.041 over the closest matching synth-only baseline. 
The pseudo recipe at \texttt{max\_len=256} underperformed its synth-only counterpart on ST1 enhanced (0.906 against 0.914, a loss of 0.008), 
which is one of the reasons we did not adopt pseudo for ST1 enhanced in the final submission.

A first diagnostic on ST3 used an NLI-style basis probe whose per-class breakdown surfaced a clear weakness: \textsf{pe} F1 was 0.25, well below the other three classes. 
That number was what motivated the \textsf{pe}-targeted synth batch above. After training the final ST3-enhanced classifier, we ran a per-class threshold sweep on the dev partition as a diagnostic; the best configuration $\{\text{ee}=0.1, \text{ei}=0.1, \text{pe}=0.05, \text{pi}=0.4\}$ added +0.026 F1 over default argmax on that partition. The submitted ST3 predictions use argmax; we did not apply these thresholds to the uploaded runs. 
A subsequent attempt to learn a basis-axis meta-learner from LLM signals gave no improvement over the encoder-only weighted average and was not used in the submission. 
\label{sec:thresholds}
The exact synthetic-pair generation prompts can be seen in Appendix~\ref{app:prompts}.

\subsection{The LLM alternative}
\label{sec:llm-detour}

Encoders did not look like the obvious choice at the start of the project. Following the published precedent of Pan et al.~\cite{pan2024zeroshotfallacy}, whose zero-shot LLM results are reported on different fallacy corpora and are therefore not directly numerically comparable, we ran a zero-shot Qwen2.5-3B baseline on ST3 enhanced, which reached F1 0.47, well below the RoBERTa baselines of \S\ref{sec:backbone}. 
A kNN few-shot variant retrieving over a \texttt{bge-large-en-v1.5} sentence embedding index (1876$\times$1024) recovered some ground at 0.53, but still trailed RoBERTa-large by a wide margin. 

We then trained CoT-LoRA recipes~\cite{hu2022lora,dettmers2023qlora,wei2022cot} on Qwen2.5-7B and Qwen2.5-32B. The 7B 5-fold CV on ST3 enhanced reached 0.85 $\pm$ 0.11; the 32B single-run reached 0.84. 
We also evaluated Qwen3-4B-Thinking~\cite{qwen3}: its 5-fold CV reached 0.96 on ST2 enhanced, which is competitive with the submitted encoder, but on ST3 base the same model reached only 0.45, below the 0.52 we already had with RoBERTa-large under the \texttt{synth\_v002} recipe. 

Despite the apparent competitiveness of the larger configurations on individual cells, the CoT-LoRA configurations we evaluated did not show a validated benefit over the encoder within our compute envelope. Our final submission (\texttt{exp\_0220}) populates all six slots with RoBERTa-large systems. 

\section{Results}
\label{sec:results}
We report both our held-out development results, on which all system-selection decisions were made, and the official Touch\'{e} 2026 test-set leaderboard results (\S\ref{sec:test}, Table~\ref{tab:test}). Development numbers are macro-averaged F1 on the held-out partition or, where flagged, the mean of a 5-fold cross-validation on that partition; all six submitted runs use recipes fixed before the test labels were released. Unless a number is explicitly labelled a test result, it is a development number.

\subsection{Validation Results}
\label{sec:headline}

Following the official Touch\'{e} 2026 release, we work from the published labelled pool for each subtask: 938 examples for ST1, the 465-example fallacious subset for ST2, and 473 non-fallacious examples for ST3. 
From each pool we hold out a stratified, roughly 15\% development split (ST1 $n=139$, ST2 $n=67$, ST3 $n=69$) on which the single-run headline numbers in Table~\ref{tab:headline} are computed, training on the remainder; the 5-fold cross-validation runs instead cross-validate over the full ST3 pool. This distinction is load-bearing for ST3: the single dev split holds exactly one \textsf{pe} example (Appendix~\ref{app:perclass}), so its macro-F1 is dragged down by that one always-hard cell, whereas 5-fold CV distributes \textsf{pe} across folds. This is the main reason the ST3-enhanced CV mean ($0.875$, \S\ref{sec:final-submission}) sits well above the single-split dev and five-seed estimates ($0.735$ and $0.720 \pm 0.015$). 
Class distributions follow the underlying Sahai et al.\ data and are long-tailed; the ST3 \textsf{pe} class in particular is small enough to be the binding constraint on macro-F1, which we revisit in \S\ref{sec:disc-pe}.

Table~\ref{tab:headline} reports precision, recall, and F1-macro on our held-out development partition for the six submitted runs. Each row carries the precision, recall, and F1-macro for one (subtask, track) cell, with macro-averaging over the 8- and 4-class label sets for ST2 and ST3 respectively; the two enhanced-track values should be read in light of the rewriter-signal discussion in \S\ref{sec:disc-leakage}. 
Across the six cells the easier binary task and the hardest scheme task differ by roughly 0.2 F1 on each track, and the base-to-enhanced gap varies markedly across subtasks.

\begin{table}[h]
\centering
\small
\caption{Headline dev results for the six submitted runs. P, R, F1 are macro-averaged
for ST2 and ST3. Precision and recall are omitted for ST3 base
(\texttt{exp\_0069}), whose saved evaluation row has a corrupted precision
column; its F1-macro is computed independently and is reliable. The two
enhanced-track values should be read with the caveat of
\S\ref{sec:disc-leakage} on the enhanced-track signal.}
\label{tab:headline}
\begin{tabular}{llcccl}
\toprule
Subtask & Track & Precision & Recall & F1 & exp\_id \\
\midrule
ST1 & base     & 0.769 & 0.762 & 0.761 & exp\_0161 \\ 
ST1 & enhanced & 0.917 & 0.914 & 0.914 & exp\_0035 \\ 
ST2 & base     & 0.746 & 0.728 & 0.730 & exp\_0032 \\ 
ST2 & enhanced & 0.970 & 0.972 & 0.970 & exp\_0072 \\ 
ST3 & base     & ---   & ---   & 0.523 & exp\_0069 \\ 
ST3 & enhanced & 0.732 & 0.740 & 0.735 & exp\_0073 \\ 
\bottomrule
\end{tabular}
\end{table}

On ST3 we additionally ran 5-fold cross-validation under our submitted recipe (fold seeds 42--46), which gave mean F1 $0.875 \pm 0.083$ (per-fold $0.77$--$0.99$) with the same configuration. 
The Qwen2.5-7B CoT-LoRA alternative for ST3 enhanced gave 0.851 $\pm$ 0.114 over the same fold split. The standard deviation overlaps the RoBERTa CV mean, which is the decision-relevant fact behind the final-submission rationale we discuss below. 
Of the six numbers, the ST2-enhanced cell is simultaneously the numerically strongest result in this paper and the one in which we have the least interpretive confidence. We address this directly in \S\ref{sec:disc-leakage}.

\subsection{Official test-set results}
\label{sec:test}

Table~\ref{tab:test} reports the official Touch\'{e} 2026 test-set leaderboard scores for our six submitted runs, together with our placement among participating teams on each (subtask, track) cell. These are the final evaluation numbers; the recipes were fixed on the development evidence above before the test labels were released.

\begin{table}[h]
\centering
\small
\caption{Official Touch\'{e} 2026 test-set results for the six submitted runs
(macro-averaged precision, recall, F1). \emph{Rank} is our placement among
participating teams on that (subtask, track) cell. Bold marks the two cells we
ranked first on.}
\label{tab:test}
\begin{tabular}{llcccc}
\toprule
Subtask & Track & Precision & Recall & F1 & Rank \\
\midrule
ST1 & base     & 0.770 & 0.754 & 0.751          & 2 \\
ST1 & enhanced & 0.946 & 0.945 & 0.944          & 2 \\
ST2 & base     & 0.876 & 0.860 & \textbf{0.859} & \textbf{1} \\
ST2 & enhanced & 0.958 & 0.954 & 0.954          & 6 \\
ST3 & base     & 0.710 & 0.724 & \textbf{0.716} & \textbf{1} \\
ST3 & enhanced & 0.716 & 0.933 & 0.748          & 5 \\
\bottomrule
\end{tabular}
\end{table}

Our uniform encoder pipeline ranked \emph{first} on both base-track classification cells --- ST2 base (0.859) and ST3 base (0.716) --- and \emph{second} on both binary-detection tracks (ST1 base 0.751, ST1 enhanced 0.944). On the two enhanced-track classification cells it placed lower (ST2 enhanced 0.954, sixth; ST3 enhanced 0.748, fifth), where the leading systems were a cascaded pipeline (\textsf{uzh-tt}, 0.991 on ST2 enhanced) and a multitask DeBERTa (\textsf{sevitha}, 0.976 on ST3 enhanced).

Two patterns are worth drawing out, both specific to the classification cells (ST2, ST3); the binary ST1 cells ranked second on both tracks and moved little between dev and test (0.761 vs.\ 0.751 base, 0.914 vs.\ 0.944 enhanced). First, the two base-track classification cells came in well above their development estimates: ST3 base rose from a dev 0.523 to a test 0.716, and ST2 base from 0.730 to 0.859, while the enhanced-track classification cells tracked their dev numbers closely (ST3 enhanced 0.735 dev vs.\ 0.748 test; ST2 enhanced 0.970 vs.\ 0.954). This is consistent with the small held-out dev splits (\S\ref{sec:headline}) underestimating base-track classification performance rather than with any test-time surprise. Second, the single-encoder recipe was most competitive on the base-track classification cells, where the enhanced rewrite is unavailable, while the enhanced-track classification cells were topped by systems built to exploit the rewritten fields, by cascading or multitasking over them; this is the same label-bearing signal we characterise in \S\ref{sec:disc-leakage}, seen from the leaderboard side.

\subsection{Binary Fallacy Detection (ST1)}
\label{sec:results-st1}

The binary task was the easiest of the three. The submitted ST1-base run (\texttt{exp\_0161}) concatenates \texttt{text\_raw} with \texttt{text\_base} at \texttt{max\_len=512} and reached F1 0.761 on the dev partition. The synthetic-data contribution is isolated separately in Table~\ref{tab:ablation} at \texttt{max\_len=256}, where +\texttt{synth\_v002} at \texttt{sw=0.5} lifts F1 from a real-only baseline of 0.716 to 0.734 (\texttt{exp\_0033}). 
On the enhanced track, the same synth recipe reached 0.914 against a real-only enhanced baseline of 0.899, with the +synth lift contributing +0.015 over that baseline. 
A pseudo-labelling extension at \texttt{max\_len=256} and \texttt{sw=0.5} underperformed its synth-only counterpart by 0.008 (Table~\ref{tab:ablation}, row $-$0.008), which is the reason ST1 enhanced ships under synth only rather than synth + pseudo. 
The 0.15 base-to-enhanced gap on ST1 previews the enhanced-track signal discussion of \S\ref{sec:disc-leakage}.

\subsection{Eight-Way Fallacy Classification (ST2)}
\label{sec:results-st2}

Fallacy classification (ST2) is comparably easy on the enhanced track (0.97) and substantially harder on the base track (0.73), the largest base-to-enhanced gap in Table~\ref{tab:headline}. On the base track the submitted +synth configuration reached 0.730 over a real-only baseline of 0.711 (+0.019). 
On the enhanced track, the +synth configuration alone reached 0.937 against a real-only enhanced baseline of 0.923 (+0.014), and the +synth+pseudo recipe at \texttt{max\_len=384} and \texttt{sw=0.3} added another +0.033 on top of that to reach 0.970. 
Per-class precision and recall for ST2-enhanced are reported in Appendix~\ref{app:perclass}; on the $n=67$ dev split the eight classes range from 0.889 to 1.000 F1, with five of the eight at 1.000 --- a high, tight distribution consistent with the rewriter carrying strong label-bearing signal as discussed in \S\ref{sec:disc-leakage}.

\subsection{Four-Way Argument Schemes (ST3)}
\label{sec:results-st3}

Argument-scheme classification (ST3) was the hardest on both tracks, at 0.523 base and 0.735 enhanced; the four-way scheme task absorbed the bulk of our experimental effort and is the cell where the per-class behaviour discussed in \S\ref{sec:disc-pe} matters most. On the base track, the +\texttt{synth\_v002} recipe at \texttt{max\_len=384} lifted F1 from a real-only baseline of 0.436 to 0.523 (+0.087, the largest single ablation gain in the paper). 
On the enhanced track, the +synth+pseudo recipe at the same window (with the augmented streams left at full weight, \texttt{sw=1.0}) reached 0.735 against a real-only enhanced baseline of 0.694 (+0.041); +synth alone tied the real-only baseline on this cell, so the gain only appeared once pseudo-labelling was added. 
Two ST3-base recipes lost ground in the ablation: the \texttt{synth\_v002 + v003} stack at \texttt{max\_len=384} dropped to 0.478 ($-$0.045 against \texttt{synth\_v002} alone), and the focal-loss variant reached 0.505 ($-$0.018 against the same baseline). 

\begin{table}[h]
\centering
\small
\caption{Per-subtask augmentation ablation, F1-macro on dev. $\Delta$ is the change
relative to the immediately preceding row within the same (subtask, track) block.
The shorthand \texttt{@N} denotes \texttt{max\_len=N} (tokeniser truncation length).
Negative deltas are not hidden.\textsuperscript{a}Single-seed unless noted; see \S\ref{sec:limitations} for the seed-variance caveat.}
\label{tab:ablation}
\begin{tabular}{llllcr}
\toprule
Subtask & Track & Recipe & exp\_id & F1 & $\Delta$ \\
\midrule
ST1 & base     & real only                          & exp\_0013 & 0.716 & ---     \\ 
ST1 & base     & + synth (sw=0.5)                   & exp\_0033 & 0.734 & +0.018 \\ 
\addlinespace
ST1 & enhanced & real only                          & exp\_0022 & 0.899 & ---     \\ 
ST1 & enhanced & + synth (sw=0.5)                   & exp\_0035 & 0.914 & +0.015 \\ 
ST1 & enhanced & + synth + pseudo @256 (sw=0.5)     & exp\_0071 & 0.906 & $-$0.008 \\ 
\midrule
ST2 & base     & real only                          & exp\_0014 & 0.711 & ---     \\ 
ST2 & base     & + synth (sw=0.5)                   & exp\_0032 & 0.730 & +0.019 \\ 
\addlinespace
ST2 & enhanced & real only                          & exp\_0023 & 0.923 & ---     \\ 
ST2 & enhanced & + synth (sw=0.5)                   & exp\_0036 & 0.937 & +0.014 \\ 
ST2 & enhanced & + synth + pseudo @384 (sw=0.3)     & exp\_0072 & 0.970 & +0.033 \\ 
\midrule
ST3 & base     & real only                          & exp\_0058 & 0.436 & ---     \\ 
ST3 & base     & + synth\_v002 @384                 & exp\_0069 & 0.523 & +0.087 \\ 
ST3 & base     & + synth\_v002 + v003 @384          & exp\_0068 & 0.478 & $-$0.045 \\ 
ST3 & base     & + synth\_v002 + focal loss         & exp\_0053 & 0.505 & $-$0.018 \\ 
\addlinespace
ST3 & enhanced & real only                          & exp\_0024 & 0.694 & ---     \\ 
ST3 & enhanced & + synth @384                       & exp\_0067 & 0.694 & 0.000  \\ 
ST3 & enhanced & + synth + pseudo @384 (sw=1.0)     & exp\_0073 & 0.735 & +0.041 \\ 
\bottomrule
\end{tabular}
\end{table}

Augmentation strategies are recipe-sensitive: synth alone gives small gains on ST1/ST2 (the smallest within the measured seed variance) and the largest single gain on ST3 base, while pseudo-labelling unlocks the enhanced track for ST2 and ST3 but costs on ST1 enhanced under the pseudo recipe. The \texttt{synth\_v003} top-up on ST3 base and the focal-loss recipe on ST3 base illustrate the same lesson at smaller scale. We keep the losses in the same table rather than relegating them to a footnote.

Table~\ref{tab:backbone} reports the matched ST3-enhanced backbone comparison, and Table~\ref{tab:backbone-other} the off-track context rows (DeBERTa on ST3 base, and the Qwen3-4B-Thinking comparator on ST2 enhanced and ST3 base) that are \emph{not} directly comparable to it. Each row is a single dev-partition F1 unless flagged as a 5-fold mean. The cross-row variability is wide enough that the comparison should be read alongside the cross-validation evidence in the prose below, not as a clean ranking.

\begin{table}[h]
\centering
\small
\caption{Matched backbone comparison on ST3 enhanced. Single-run dev F1 unless
marked as a 5-fold cross-validation mean$\pm$sd. The submitted row is marked
$\dagger$ (its 5-fold CV mean is given alongside the single-run number).}
\label{tab:backbone}
\begin{tabular}{llc}
\toprule
Backbone & exp\_id & F1 (ST3 enh) \\
\midrule
RoBERTa-base                      & exp\_0018 & 0.675 \\ 
RoBERTa-large$^{\dagger}$         & exp\_0073 & 0.735 \; (CV $0.875 \pm 0.083$) \\ 
ModernBERT-large                  & exp\_0086 & 0.719 \\ 
Qwen2.5-3B zero-shot              & exp\_0091 & 0.467 \\ 
Qwen2.5-3B kNN few-shot           & exp\_0143 & 0.529 \\ 
Qwen2.5-7B CoT-LoRA (5-fold CV)   & exp\_0180 & $0.851 \pm 0.114$ \\ 
Qwen2.5-32B CoT-LoRA (single)     & exp\_0207 & 0.841 \\ 
\bottomrule
\end{tabular}
\end{table}

\begin{table}[h]
\centering
\small
\caption{Off-track context rows, reported for completeness and \emph{not}
comparable to the ST3-enhanced column of Table~\ref{tab:backbone}: DeBERTa on
ST3 base (the large collapsed under NaN, the base merely underperformed;
\S\ref{sec:backbone}) and the Qwen3-4B-Thinking comparator on ST2 enhanced and
ST3 base.}
\label{tab:backbone-other}
\begin{tabular}{lllc}
\toprule
Backbone & Subtask & Track & F1 \\
\midrule
DeBERTa-v3-base                & ST3 & base & 0.277 \\ 
DeBERTa-v3-large (collapsed)   & ST3 & base & 0.205 \\ 
Qwen3-4B-Thinking (5-fold CV)  & ST2 & enh  & 0.963 \\ 
Qwen3-4B-Thinking (5-fold CV)  & ST3 & base & 0.448 \\ 
\bottomrule
\end{tabular}
\end{table}

Within the matched ST3-enhanced comparison (Table~\ref{tab:backbone}), RoBERTa-large is the submitted system and RoBERTa-base sits 0.06 F1 below it, which is the empirical justification for paying the larger backbone's compute cost. ModernBERT-large at 0.719 is the next-best encoder route but loses 0.016 F1 on this cell. The Qwen rows tell their own story: the 3B zero-shot configuration trails RoBERTa-large by 0.27 F1, and kNN few-shot recovers some ground but still trails by 0.21. The fine-tuned 7B and 32B CoT-LoRA configurations close most of the gap on a single run, but under cross-validation the encoder and the LLM are not separable: the encoder's 5-fold mean of $0.875 \pm 0.083$ and the Qwen2.5-7B mean of $0.851 \pm 0.114$ have heavily overlapping standard-deviation bands, so the comparison is statistically inconclusive at our sample size. The off-track rows (Table~\ref{tab:backbone-other}) are context only: DeBERTa-v3 did not train usefully under any recipe we tried (0.277 on ST3 base, and 0.205 with the NaN-explosion history we detail in \S\ref{sec:backbone}), so its low numbers reflect a training failure rather than a track-matched loss, while the Qwen3-4B-Thinking comparator is competitive on ST2 enhanced under cross-validation (0.963) but below the RoBERTa baseline on ST3 base (0.448), which complicates a clean encoder-beats-LLM headline.

The per-class F1 distribution for the submitted ST3-enhanced system is reported in Appendix~\ref{app:perclass} (\textsf{ee} 1.000, \textsf{ei} 0.979, \textsf{pi} 0.963, \textsf{pe} 0.000), computed under argmax on the $n=69$ dev split, and discussed in \S\ref{sec:disc-pe}. The \textsf{pe} cell rests on a single dev example and is therefore uninformative; we rely on the 5-fold cross-validation (\S\ref{sec:disc-pe}) for any \textsf{pe} characterisation. The submitted run's dev logits were not checkpointed, so its exact confusion matrix cannot be recovered. We have since added dev-logit checkpointing to the training script and re-ran the \texttt{exp\_0073} recipe under five seeds on the same fixed dev split, both to characterise run-to-run variance and to recover a representative confusion matrix (Table~\ref{tab:st3cm}). Across the five re-runs dev macro-F1 is $0.720 \pm 0.015$ (range $0.700$--$0.741$, all fifteen epochs completed), which places the submitted $0.735$ near the top of the observed range and is consistent with the single-run caveat of \S\ref{sec:limitations}. The per-class pattern is stable across seeds: \textsf{ei} ($0.98$--$1.00$) and \textsf{pi} ($0.963$ in every run) are recovered almost perfectly, the variance concentrates in the seven-example \textsf{ee} class ($0.86$--$1.00$), and \textsf{pe} has recall $0$ in all five re-runs. 
We turn to the interpretation of the enhanced-track numbers and the rare-class behaviour we observed on ST3 in \S\ref{sec:discussion}.

\subsection{Final submission rationale}
\label{sec:final-submission}

We did not use the additional two-runs-per-track budget available under the Touch\'{e} submission rules. We report on the single CV-validated configuration per slot. The non-obvious choice was ST3 enhanced, where the Qwen2.5-32B single-run F1 of 0.841 sat above the RoBERTa-large dev number of 0.735 but just below the RoBERTa-large 5-fold CV mean of 0.875. 
Two reasons drove the encoder choice, and we keep them separate. The first is operational: the Qwen2.5-32B 5-fold cross-validation was still running at the deadline, leaving the RoBERTa-large CV as the only completed cross-validation we had in hand at submission time. The second is the absence of a statistical reason to prefer Qwen even if the comparison had been completed: a 5-fold standard deviation of $\pm$0.114 on Qwen2.5-7B is evidence the encoder-vs-LLM comparison is underpowered at this sample size, not evidence for the system whose CV mean is higher. 
We therefore do not claim the available CV evidence rules out the Qwen alternatives; we only claim that at deadline the encoder pipeline was the only fully-validated configuration. A post-hoc ensemble of the available encoder and LLM systems gave no gain over the single-source encoder, which is one further reason we did not retroactively complicate the submission. 
A submission dry-run verified that the TIRA-format outputs matched the lab's expected schema before the final upload. 

\section{Discussion and Future Work}
\label{sec:discussion}
This year we reflect on the structural shape of the Touch\'{e} 2026 evaluation and on what our submissions can and cannot say about it. The first three subsections below close the four contributions stated in \S\ref{sec:intro}: \S\ref{sec:disc-encoder} closes C3 (encoder vs.\ LLM at this scale), \S\ref{sec:disc-leakage} closes the first half of C4 (the enhanced-track signal), and \S\ref{sec:disc-pe} closes the second half of C4 (the \textsf{pe} rare-class limitation). \S\ref{sec:limitations} states the limitations attached to the numbers behind those subsections, and \S\ref{sec:future} lists the most useful follow-ups the present work allows.

\subsection{Encoder vs.\ LLM at this scale}
\label{sec:disc-encoder}

Across the cells we were able to validate under cross-validation, encoder fine-tuning matched or edged out every LLM track we evaluated on point estimates. The most informative comparison is on ST3 enhanced, where the encoder's 5-fold CV mean of $0.875 \pm 0.083$ and the Qwen2.5-7B CoT-LoRA comparator's $0.851 \pm 0.114$ have standard-deviation bands that overlap almost entirely. The single-run Qwen2.5-32B reached 0.841 on the same task. 
With dispersion that wide on both sides, the apparent 0.024 gap is not statistically meaningful from a single 5-fold split. The strongest claim the evidence supports is that the LLM track did not show a validated benefit at this dataset scale within our compute envelope, not that LLMs are categorically worse. We scope the claim to this dataset (roughly 473 ST3 examples), this label structure, and our compute budget; we would not extrapolate to larger labelled corpora or to richer reasoning targets. The test outcome is consistent with this reading: the encoder pipeline ranked first on both base-track classification cells (ST2 and ST3, \S\ref{sec:test}), where no rewritten fields are available, while the enhanced-track cells were topped by cascaded and multitask systems built to exploit the rewrite --- an architecture axis, not a backbone-family verdict.

\subsection{What the enhanced-track signal contains}
\label{sec:disc-leakage}

The 0.970 ST2-enhanced and 0.914 ST1-enhanced numbers are the strongest in the paper and warrant a careful read of what the rewritten fields contribute. The lab provides the enhanced track as a designed difficulty setting: \texttt{text\_enhanced} and \texttt{argument\_enhanced} are organiser-supplied paraphrases produced with knowledge of the gold labels, and base-only systems are scored in a separate track~\cite{kiesel:2026c}. The legitimate scientific question is therefore not whether something illicit happened (it did not) but how much of the enhanced-track lift reflects the rewrite carrying label-bearing signal versus genuine argument clarification. We address the question in four steps. 

\emph{(a) The setup.} Any classifier trained on \texttt{text\_enhanced} or \texttt{argument\_enhanced} has access to features that a base-only pipeline would have to recover from raw text. The interesting question is not whether such features exist (they are designed in) but whether the classifier is exploiting them as label strings or as cleaner argumentative structure.

\emph{(b) A sanity check, and what it does not establish.} We ran a regex-mask sanity check that replaced known fallacy-name strings and obvious cue words in the enhanced fields and re-trained the ST2-enhanced classifier on the masked inputs. F1 was retained at 0.937 after masking, against the unmasked enhanced-track \emph{baseline} (no pseudo-label) of 0.937. The mask was applied to this synth-only configuration, not to the submitted \texttt{+synth+pseudo} system (0.970), so it does not bear on any signal the pseudo-label stage might add. 
The mask deliberately covers the eight ST2 fallacy-class names and their morphological variants only, not the full semantic content of the rewritten fields. A null delta is therefore the expected outcome \emph{if} the classifier was not relying on those literal strings as a shortcut, and is the specific hypothesis the check rules out, not a general claim that the rewritten fields contain no label-bearing signal at all. We call this a sanity check rather than a diagnostic for exactly this reason: it tests a hypothesis the rewriter's design made a priori unlikely, and therefore says little about the softer semantic channel.

\emph{(c) The interpretation.} A retained F1 of 0.937 under aggressive string masking means that trivial string-matching of fallacy labels in \texttt{text\_enhanced} is not the dominant signal the classifier relies on. That rules out the most superficial story, but the rewriter could still encode the label in word choice, sentence structure, hedging, or topical framing without ever using a fallacy-name string. We note that the ST2-enhanced per-class F1 values (Appendix~\ref{app:perclass}) range from 0.889 to 1.000 on the $n=67$ dev split, with five of the eight classes at 1.000 --- a high, tight distribution still consistent with strong label-bearing signal in the rewritten fields, though the wider spread than we first reported tempers that reading slightly. 

\emph{(d) The reporting commitment.} We present base-track and enhanced-track numbers separately throughout this notebook and do not aggregate them into a single headline; where the paper takes a position e.g.,\ on the ablation pattern, on the encoder vs.\ LLM comparison, on the \textsf{pe} limitation, we anchor that position on base-track behaviour or 5-fold CV numbers rather than on enhanced-track point estimates. The headline enhanced-track numbers stand as ceiling results for the designed track they were produced under.

The dev-overlap audit of \S\ref{sec:pseudo} addresses a different worry: that the pseudo-labelled examples, drawn from the unlabelled pool, might be near-duplicates of dev instances and so inflate the dev score directly; it found no such overlap. 
It does not, however, speak to whether the rewritten fields carry label-bearing signal --- that is what the masking check above probes, and neither check clears the softer semantic channel. A falsifiable extension, training a classifier on base-only fields and evaluating it on enhanced fields, would isolate the marginal contribution of the rewriter directly; we flag it in \S\ref{sec:future}.

\subsection{The \textsf{pe} rare-class limitation}
\label{sec:disc-pe}

The practical-external (\textsf{pe}) class on ST3 was the hardest scheme class in every evaluation that contained enough \textsf{pe} examples to measure it. An NLI-style basis probe flagged it early. 
We responded with three interventions: a \textsf{pe}-targeted synthetic batch, a per-class threshold sweep that pushed the \textsf{pe} threshold to 0.05, and an LLM basis-axis meta-learner; none produced a measurable lift on the dev split, which holds too few \textsf{pe} examples to detect one. 

Quantifying the gap is constrained by how few \textsf{pe} examples the dev partition holds. Our single held-out dev split contains exactly one \textsf{pe} example, so the submitted-run \textsf{pe} numbers --- F1 $=0.000$ under argmax (\texttt{exp\_0073}) and $0.25$ under the dev-only threshold sweep (\texttt{exp\_0145}), both computed on that same one-example split --- reflect a single instance rather than a distribution and cannot support a ceiling estimate. The only evaluation with meaningful \textsf{pe} support is the 5-fold cross-validation under the submitted recipe (\texttt{exp\_0153}), where \textsf{pe}-F1 averages $0.713$ across folds against a $0.875$ macro-F1, but with wide fold-to-fold variance (per-fold $0.40$--$1.00$, standard deviation $0.25$). 
Even there, \textsf{pe} is consistently the class that most depresses macro-F1, and the variance underlines how few \textsf{pe} instances the corpus provides; the thin \textsf{pe} representation in the dev partition is itself a limitation of the evaluation (\S\ref{sec:limitations}). The substantive finding stands: \textsf{pe} is the rare class that the targeted synthetic augmentation, the per-class threshold sweep, and the basis-axis meta-learner all failed to lift, and targeted \textsf{pe}-class data collection (\S\ref{sec:future}) is the intervention most likely to help. A five-seed re-run of the submitted recipe (Table~\ref{tab:st3cm}, \S\ref{sec:results-st3}) reinforces the point from the other direction: \textsf{ei} and \textsf{pi} are recovered almost perfectly in every run while the single \textsf{pe} dev example is misclassified in all five, so its dev F1 of $0$ reflects the split's inability to measure the class rather than a stable property of the system.

\subsection{Limitations}
\label{sec:limitations}

Several limitations should accompany the numbers in this notebook.

System selection was made entirely on our held-out dev partition and 5-fold cross-validation; the official test results (Table~\ref{tab:test}, \S\ref{sec:test}) became available only afterwards and informed no decision in this paper. Where a single 5-fold split was available (ST3 enhanced under our submitted recipe, the Qwen2.5-7B CoT-LoRA comparator, and the Qwen3-4B-Thinking comparators) we use the CV mean as the decision-relevant estimate; the other dev numbers in Tables~\ref{tab:headline}--\ref{tab:backbone} are single-run point estimates, and the ranking of close-together rows should be read with appropriate caution. 

We ran a five-seed variance check only on the submitted ST3-enhanced cell (\S\ref{sec:results-st3}, dev macro-F1 $0.720 \pm 0.015$); we did not repeat it for the other five cells, whose reported single-run dev numbers should be read as point estimates. 
For runs whose individual seeds are not recorded, we used the lab-default \texttt{seed=42}; CV runs used fold seeds 42--46.

The enhanced-track headline numbers are upper bounds on what the designed track can yield, in the sense developed in \S\ref{sec:disc-leakage}. The regex-mask sanity check rules out the trivial-string channel; it does not clear the softer semantic channel, and we have flagged this everywhere the enhanced numbers appear. 

The per-class F1 breakdown in Appendix~\ref{app:perclass} is computed under argmax decoding for both ST2 and ST3, matching the decoding used for the submitted runs and the headline numbers in Table~\ref{tab:headline}. The ST3 \textsf{pe} cell in particular rests on a single dev example and should not be read as a per-class estimate; the 5-fold cross-validation (\S\ref{sec:disc-pe}) is the only evaluation with meaningful \textsf{pe} support. 

The per-class analysis in this version of the notebook (Appendix~\ref{app:perclass} and \S\ref{sec:disc-pe}) differs from the initially submitted version. The submitted version reported a per-class breakdown whose values and supports did not trace to any saved evaluation artefact; during code--paper auditing we reconciled the analysis against the actually-submitted checkpoints (\texttt{exp\_0072}, \texttt{exp\_0073}) and now report the reproducible argmax numbers on the real dev partitions ($n=67$ and $n=69$).

Overall, these limitations do not invalidate the above numbers, but the enhanced-track results in particular should be read as upper bounds on what a fair pipeline would extract from the same inputs.

\subsection{Future work}
\label{sec:future}

The most useful follow-ups the present work allows are concrete. A falsifiable extension of the leakage diagnostic, training a classifier on base-only fields and evaluating it on enhanced fields, would isolate the marginal contribution of the rewriter and let us put a number on the residual semantic-leakage channel; we did not run this within our compute envelope and flag it as the most useful follow-up the C4 thread allows. A leakage-controlled re-evaluation on a future Touch\'{e} edition, one that ships a separate decoder for the rewritten fields, or that withholds the label-aware rewrite from training entirely, would let us decide more cleanly between the encoder and LLM tracks at this dataset scale. A larger compute budget on the Qwen2.5-32B fold split would let us close the comparison we left underpowered, addressing the open question behind C3. A per-class threshold sweep on ST3 base, paralleling the sweep we ran on ST3 enhanced, would let us test whether the \textsf{pe} limitation we observed is a property of the enhanced track or of the four-way label structure more broadly. And targeted \textsf{pe}-class data collection, rather than further synthetic augmentation, looks like the most useful intervention if the rare-class limitation is to be overcome at all.

\section{Conclusion}
\label{sec:conclusion}
We close on the four contributions in the same order they appear in \S\ref{sec:intro}. \textbf{C1:} A single RoBERTa-large fine-tuning pipeline, with only per-cell variation in input fields and loss weights, populated all six (subtask, track) cells, ranking first on the official test set for both base-track classification cells (ST2 and ST3) and second on both binary-detection tracks (\S\ref{sec:test}). \textbf{C2:} Synthetic contrastive augmentation gave small gains (+0.014 to +0.087 across cells, the smallest of which fall within the $\pm 0.015$ seed variance we measured), and pseudo-labelling unlocked the enhanced track for ST2 and ST3, but a pseudo recipe lost ground on ST1 enhanced and a rare-class synthetic top-up lost ground on ST3 base; we reported both kinds of result in the same tables. \textbf{C3:} A grid covering DeBERTa-v3, ModernBERT, and four Qwen variants up to 32B parameters showed no validated benefit over the encoder at this corpus scale; the 5-fold standard deviation on the strongest LLM comparator leaves that comparison underpowered rather than closed. \textbf{C4:} The enhanced-track headline numbers should be read as upper bounds because the rewritten fields carry signal that our string-mask sanity check could not rule out at the semantic level, and the practical-external (\textsf{pe}) ST3 class continued to limit macro-F1 after targeted augmentation, threshold calibration, and a basis-axis meta-learner. With its 2026 addition of fallacy detection, Touch\'{e} gives argument-aware NLP a benchmark we expect to return to, and the negative results collected here are offered to save the next team the detours we took.

\begin{acknowledgments}
The author thanks the Touch\'{e} 2026 lab organizers for providing
the dataset, evaluation infrastructure, and submission platform.
\end{acknowledgments}

\section*{AI Usage Declaration}
This text was proofread and improved with the assistance of
Claude (Anthropic). The following prompt was used:

\begin{quote}
``Imagine you're an AI master student. Evaluate my text based on the
task description and check if it correct or not, if not tell me which
parts I should improve.''
\end{quote}

\noindent Claude was used to evaluate the text against the
Touch\'{e} 2026 Fallacy Detection task description, identify missing
elements (results coverage, leakage handling, methodology and ablation
details) and suggest phrasing improvements. The author conceived and
executed all experiments, made all submission decisions, and takes full
responsibility for the publication's content.

\bibliography{references}

\appendix

\section{Prompts}
\label{app:prompts}

Synthetic contrastive pairs and pseudo-labelled examples contributed to
the submitted systems (\S\ref{sec:synth}, \S\ref{sec:pseudo}). The system
prompts that drove each generation batch are reproduced verbatim below;
the matching user prompts are built per call from a topic seed, a
subreddit, a small block of real exemplars from the train split, a
sampled target length and number of supports, and (for ST1/ST2) a
fallacy card or (for ST3) a scheme pair. Generated batches are released
alongside this notebook in the TIRA submission package.

\medskip

\begin{promptbox}{Prompt 1: ST1/ST2 contrastive-pair generation.
System message.}
You generate training data for a fallacy detection model. The data
must read as authentic Reddit replies: informal register, hedges
(``idk'', ``tbh'', ``iirc''), occasional typos, no debate-club
phrasing, no signposting. Never name the fallacy inside the
generated text.

The `load-bearing signals' you'll be shown describe the LOGICAL
PATTERN of the fallacy, not phrases to use verbatim. Express the
same reasoning move with fresh wording. Do NOT use any of these
stock formulations (fresh wording only): `millions of people
can't be wrong', `people can't be wrong', `everyone knows',
`no middle ground', `no in between', `before you know it',
`next thing you know', `where does it end', `it's a slippery
slope', `we've always done it', `natural is better', `chemicals
are bad', `yeah i agree', `i agree it's'. If your draft contains
any of these phrases, rewrite the draft.

You will produce a CONTRASTIVE PAIR: one reply that commits the
target fallacy, and one reply using structurally similar reasoning
that does NOT commit it. Both replies must address the same topic
with claims of similar polarity, so the only meaningful difference
is the soundness of the inference.

CRITICAL: Do NOT cite specific percentages, year-tagged studies
(e.g.\ `2023 study'), polls, named institutions (Gallup, Pew, CDC,
etc.), or made-up statistics in EITHER reply. Both replies must
reason from lived experience, anecdote, observation, or general
claims. The difference between them is the soundness of the
inference, not the presence of evidence.
\end{promptbox}

The user prompt supplies a fallacy card (definition, load-bearing
signals, legitimate counterpart), the topic and subreddit, a target
length and per-reply support count, and a small block of real
exemplars. The model returns a JSON object carrying \texttt{thread\_title},
\texttt{parent}, and two replies (\texttt{fallacious} and
\texttt{legitimate}), each with \texttt{text\_raw}, \texttt{text\_base},
\texttt{claim}, and \texttt{supports}.

\medskip

Prompt 2 has the same shape as Prompt 1, but produces two
non-fallacious replies using different argumentation schemes.

\begin{promptbox}{Prompt 2: ST3 contrastive-pair generation
(\texttt{synth\_v002} ST3). System message.}
You generate training data for an argumentation scheme classifier.
The data must read as authentic Reddit replies: informal register,
hedges, occasional typos, natural conversation. No academic phrasing.

[blocklist as in Prompt 1]

You will produce a CONTRASTIVE PAIR: two non-fallacious replies to
the same parent comment, on the same topic, with similar conclusions
-- but using DIFFERENT argumentation schemes. The structural
difference should be clear but natural.
\end{promptbox}

The user prompt specifies Scheme A (target) and Scheme B (contrast),
each with goal axis (practical / epistemic), basis axis (internal /
external), definition, and load-bearing signals, together with the
topic, subreddit, target length, and exemplar block. JSON output is
\texttt{scheme\_a} / \texttt{scheme\_b}, each with \texttt{scheme},
\texttt{argument\_goal}, \texttt{argument\_basis}, \texttt{text\_raw},
\texttt{text\_base}, \texttt{claim}, and \texttt{supports}.

\medskip

A separate, simpler prompt generated the \textsf{pe}-targeted batch
(50 examples per label, \texttt{openai/gpt-4o-mini}, temperature 0.9).

\begin{promptbox}{Prompt 3: \textsf{pe}-targeted generation
(\texttt{exp\_0103}). System message.}
You are an expert in argumentation theory and propaganda techniques.
Generate realistic argumentative text snippets that contain propaganda
or emotional manipulation techniques (loaded language, fear appeals,
bandwagon, false dilemma, etc.) rather than logical reasoning.

The text should be 2-4 sentences, resemble real-world political
discourse or persuasive media, and clearly exemplify propaganda /
emotional fallacies.
\end{promptbox}

\medskip

\noindent Two user prompts, one per generated label:

\begin{promptbox}{Prompt 3: \textsf{pe}-targeted generation
(\texttt{exp\_0103}). User prompts.}
\textbf{[pe]}\quad Generate a short argumentative text (2-4 sentences)
that uses propaganda or emotional manipulation. Include at least one
of: loaded language, appeal to fear, bandwagon appeal, false dilemma,
or emotional blackmail. Make it sound like real political/media
discourse.

\textbf{[pa]}\quad Generate a short argumentative text (2-4 sentences)
that primarily appeals to emotion (pathos) rather than logic. Use
emotional language, fear, pity, or desire to persuade. Make it
realistic.
\end{promptbox}

\paragraph{Pseudo-label confidence selection (\texttt{exp\_0072},
\texttt{exp\_0073}).}
The pseudo-labelling step filtered the encoder classifier's own
predictions at a nominal $\tau=0.9$, relaxed for under-populated classes
so that rare schemes were retained (see \S\ref{sec:pseudo}); it is not LLM-API
driven and uses no prompt. We list it here for completeness rather
than because there is a prompt to record.

\section{Hyperparameter Grid}
\label{app:hyper}

Table~\ref{tab:hyper} summarises the configurations used for the six submitted runs.
All runs share the RoBERTa-large backbone~\cite{liu2019roberta} and use the
\texttt{transformers} library~\cite{wolf2020transformers}. Optimiser is AdamW with
linear warmup over the first 10\% of steps. Loss is class-weighted cross-entropy for ST2 and ST3;
ST1 uses unweighted cross-entropy. Where the experiment log does not separately record a seed,
we used the lab default \texttt{seed=42}; 5-fold CV runs used fold seeds 42--46.

\begin{table}[h]
\centering
\small
\caption{Submitted-run hyperparameter grid. \texttt{sw} is the loss weight applied to
synthetic (and pseudo-labelled) examples relative to real examples. \texttt{max\_len}
is the tokeniser truncation length.}
\label{tab:hyper}
\begin{tabular}{llcccccc}
\toprule
Subtask & Track & exp\_id & max\_len & batch & lr & epochs & extras \\
\midrule
ST1 & base     & exp\_0161 & 512 & 8 & 1e-5 & 12 & synth\_v002 sw=0.5, raw\_concat \\ 
ST1 & enhanced & exp\_0035 & 256 & 8 & 1e-5 & 12 & synth\_v002 sw=0.5 \\ 
ST2 & base     & exp\_0032 & 256 & 8 & 1e-5 & 15 & synth\_v002 sw=0.5 \\ 
ST2 & enhanced & exp\_0072 & 384 & 8 & 1e-5 & 12 & synth + pseudo sw=0.3 \\ 
ST3 & base     & exp\_0069 & 384 & 8 & 1e-5 & 15 & synth\_v002 sw=0.5 \\ 
ST3 & enhanced & exp\_0073 & 384 & 8 & 1e-5 & 15 & synth + pseudo sw=1.0 \\ 
\bottomrule
\end{tabular}
\end{table}

Synthetic generation used a frontier LLM via API (see \S\ref{sec:system});
the prompt template and the generated pairs are released alongside this
notebook in the TIRA submission package. Per-class decision thresholds for
ST3 were swept on the dev partition; the best configuration was
$\{\text{ee}{=}0.1, \text{ei}{=}0.1, \text{pe}{=}0.05, \text{pi}{=}0.4\}$
(\texttt{exp\_0145}, $+0.026$ F1 over default argmax on the dev partition; this sweep was a dev-partition diagnostic and was not applied to the submitted predictions, which use argmax). 

\section{Per-Class F1 on the Dev Partition}
\label{app:perclass}

Table~\ref{tab:perclass} reports per-class precision, recall, F1, and
support for the submitted enhanced-track systems on the held-out
development partition. Both ST2 (\texttt{exp\_0072}) and ST3
(\texttt{exp\_0073}) values are computed from default argmax predictions
--- the decoding actually used for the submitted runs --- and are read
directly from each checkpoint's saved metrics ($n=67$ for the ST2 dev
split, $n=69$ for ST3). Both macro-F1 values reconcile with
Table~\ref{tab:headline} (\S\ref{sec:results}) within $\pm 0.001$
rounding. This breakdown was reconciled against the submitted checkpoints
during code--paper auditing, replacing an earlier table whose per-class
values and supports did not correspond to any saved evaluation artefact.

\begin{table}[h]
\centering
\small
\caption{Per-class precision, recall, F1, and support on the dev
partition for ST2-enhanced (\texttt{exp\_0072}, $n=67$) and ST3-enhanced
(\texttt{exp\_0073}, $n=69$), both under argmax decoding. Values are read
from the submitted checkpoints' saved metrics. The ST3 \textsf{pe} cell
rests on a single dev example and is not an informative per-class
estimate (see \S\ref{sec:disc-pe}).}
\label{tab:perclass}
\begin{tabular}{lcccc}
\toprule
Label & Precision & Recall & F1 & Support \\
\midrule
\multicolumn{5}{l}{\textit{ST2 (8-way; exp\_0072; argmax; $n=67$)}} \\ 
authority             & 1.000 & 1.000 & 1.000 & 8 \\
black-white           & 1.000 & 1.000 & 1.000 & 8 \\
hasty-generalization  & 1.000 & 1.000 & 1.000 & 8 \\
natural               & 1.000 & 0.889 & 0.941 & 9 \\
population            & 0.875 & 1.000 & 0.933 & 7 \\
slippery-slope        & 1.000 & 1.000 & 1.000 & 9 \\
tradition             & 0.889 & 0.889 & 0.889 & 9 \\
worse-problems        & 1.000 & 1.000 & 1.000 & 9 \\
\midrule
\multicolumn{5}{l}{\textit{ST3 (4-way; exp\_0073; argmax; $n=69$)}} \\ 
ee (epistemic-external) & 1.000 & 1.000 & 1.000 & 7 \\
ei (epistemic-internal) & 1.000 & 0.958 & 0.979 & 48 \\
pi (practical-internal) & 0.929 & 1.000 & 0.963 & 13 \\
pe (practical-external) & 0.000 & 0.000 & 0.000 & 1 \\
\bottomrule
\end{tabular}
\end{table}

\section{ST3-Enhanced Confusion Matrix}
\label{app:st3cm}

Table~\ref{tab:st3cm} gives a representative dev confusion matrix for the
ST3-enhanced system. The submitted run's logits were not checkpointed, so its
exact matrix is unrecoverable; the training script now saves per-example dev
logits (\texttt{dev\_logits.npz}), and we re-ran the \texttt{exp\_0073} recipe
under five seeds on the same fixed dev split (\S\ref{sec:results-st3}). The
matrix shown is the run nearest the five-seed mean (macro-F1 $0.722$). The
qualitative structure is stable across seeds: \textsf{ei} and \textsf{pi} are
recovered almost perfectly, the residual errors sit in the small \textsf{ee}
class, and the single \textsf{pe} example is misclassified in every run
(\S\ref{sec:disc-pe}).

\begin{table}[h]
\centering
\small
\caption{Representative dev confusion matrix for ST3-enhanced (\texttt{exp\_0073}
recipe, seed nearest the five-seed mean; macro-F1 $0.722$, $n=69$). Rows are the
gold label, columns the argmax prediction. Reconstructed from the released
\texttt{dev\_logits.npz}.}
\label{tab:st3cm}
\begin{tabular}{lcccc}
\toprule
 & \multicolumn{4}{c}{Predicted} \\
\cmidrule(l){2-5}
True & \textsf{ee} & \textsf{ei} & \textsf{pi} & \textsf{pe} \\
\midrule
\textsf{ee} (epistemic-external) & 6 & 0 & 0 & 1 \\
\textsf{ei} (epistemic-internal) & 0 & 48 & 0 & 0 \\
\textsf{pi} (practical-internal) & 0 & 0 & 13 & 0 \\
\textsf{pe} (practical-external) & 0 & 0 & 1 & 0 \\
\bottomrule
\end{tabular}
\end{table}

\section{Citation Graph}
\label{app:citgraph}

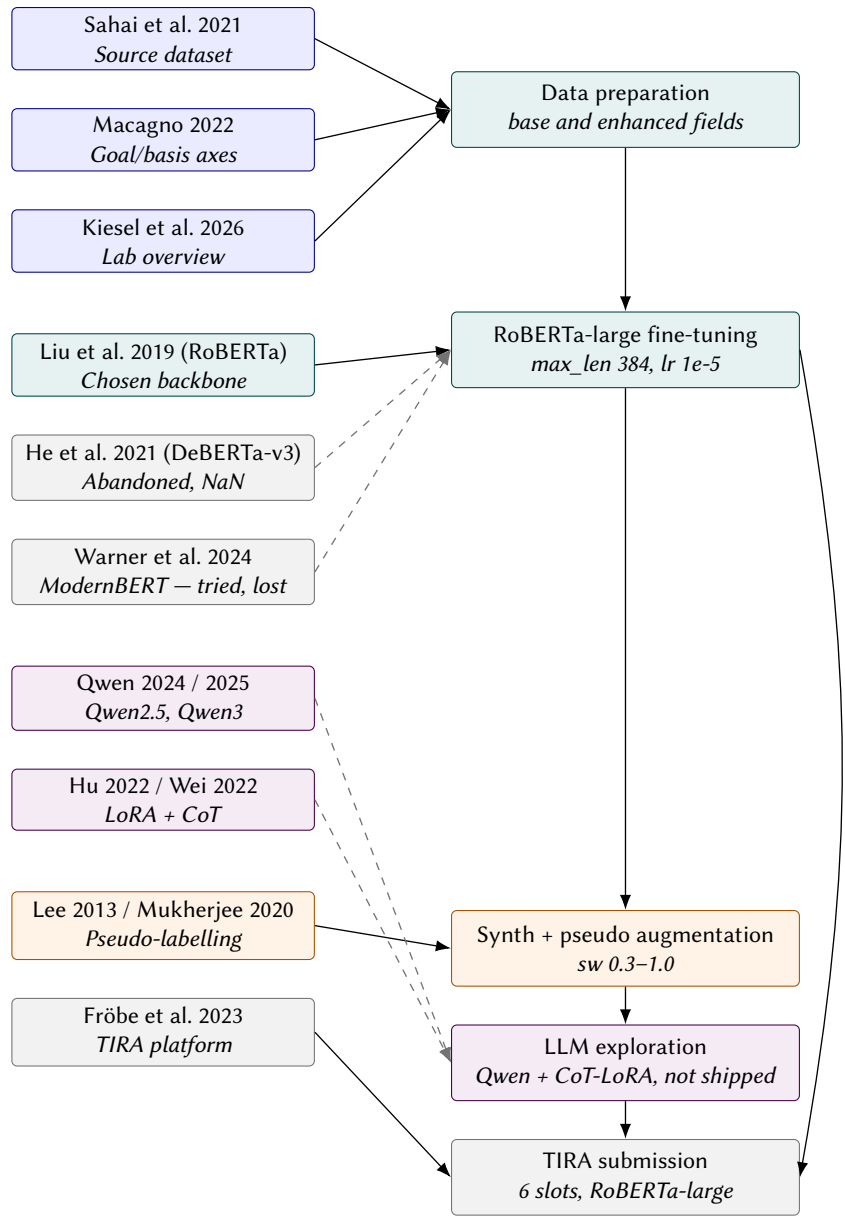
\begin{figure}[t]
\centering
\begin{tikzpicture}[
  node distance = 5mm and 18mm,
  font          = \footnotesize,
  every node/.style = {align=center},
  citebox/.style = {draw, rounded corners=2pt, inner sep=3pt,
                    minimum width=40mm, minimum height=8mm,
                    line width=0.4pt},
  pipebox/.style = {draw, rounded corners=2pt, inner sep=3pt,
                    minimum width=46mm, minimum height=10mm,
                    line width=0.4pt},
  blue/.style   = {citebox, fill=blue!8,   draw=blue!55!black},
  teal/.style   = {citebox, fill=teal!10,  draw=teal!55!black},
  gray/.style   = {citebox, fill=black!5,  draw=black!55},
  purp/.style   = {citebox, fill=violet!8, draw=violet!55!black},
  amber/.style  = {citebox, fill=orange!10,draw=orange!65!black},
  ptl/.style    = {pipebox, fill=teal!10,  draw=teal!55!black},
  pgr/.style    = {pipebox, fill=black!5,  draw=black!55},
  pam/.style    = {pipebox, fill=orange!10,draw=orange!65!black},
  ppu/.style    = {pipebox, fill=violet!8, draw=violet!55!black},
  solid/.style  = {-{Latex[length=2mm]}, line width=0.5pt},
  dropped/.style= {-{Latex[length=2mm]}, line width=0.5pt, dashed,
                   draw=black!55},
]

\node[blue]  (c1)                            {Sahai et al.\ 2021\\\textit{Source dataset}};
\node[blue,  below=of c1] (c2)               {Macagno 2022\\\textit{Goal/basis axes}};
\node[blue,  below=of c2] (c3)               {Kiesel et al.\ 2026\\\textit{Lab overview}};
\node[teal,  below=8mm of c3] (c4)           {Liu et al.\ 2019 (RoBERTa)\\\textit{Chosen backbone}};
\node[gray,  below=of c4] (c5)               {He et al.\ 2021 (DeBERTa-v3)\\\textit{Abandoned, NaN}};
\node[gray,  below=of c5] (c6)               {Warner et al.\ 2024\\\textit{ModernBERT --- tried, lost}};
\node[purp,  below=8mm of c6] (c7)           {Qwen 2024 / 2025\\\textit{Qwen2.5, Qwen3}};
\node[purp,  below=of c7] (c8)               {Hu 2022 / Wei 2022\\\textit{LoRA + CoT}};
\node[amber, below=8mm of c8] (c9)           {Lee 2013 / Mukherjee 2020\\\textit{Pseudo-labelling}};
\node[gray,  below=of c9] (c10)              {Fr\"{o}be et al.\ 2023\\\textit{TIRA platform}};

\node[ptl, right=of c2, yshift=4mm]   (p1) {Data preparation\\\textit{base and enhanced fields}};
\node[ptl, right=of c4, yshift=2mm]   (p2) {RoBERTa-large fine-tuning\\\textit{max\_len 384, lr 1e-5}};
\node[pam, right=of c9, yshift=-3mm]  (p3) {Synth + pseudo augmentation\\\textit{sw 0.3--1.0}};
\node[ppu, below=of p3]               (p4) {LLM exploration\\\textit{Qwen + CoT-LoRA, not shipped}};
\node[pgr, below=of p4]               (p5) {TIRA submission\\\textit{6 slots, RoBERTa-large}};

\draw[solid] (c1.east) -- (p1.west);
\draw[solid] (c2.east) -- (p1.west);
\draw[solid] (c3.east) -- (p1.west);
\draw[solid] (c4.east) -- (p2.west);
\draw[solid] (c9.east) -- (p3.west);
\draw[solid] (c10.east) -- (p5.west);

\draw[dropped] (c5.east) -- (p2.west);
\draw[dropped] (c6.east) -- (p2.west);
\draw[dropped] (c7.east) -- (p4.west);
\draw[dropped] (c8.east) -- (p4.west);

\draw[solid] (p1.south) -- (p2.north);
\draw[solid] (p2.south) -- (p3.north);
\draw[solid] (p3.south) -- (p4.north);
\draw[solid] (p4.south) -- (p5.north);
\draw[solid] (p2.east) to[bend left=10] (p5.east);

\end{tikzpicture}
\caption{Which prior work fed which component of our pipeline. Solid
arrows are routes that shipped; dashed arrows are routes we tried
and dropped (DeBERTa-v3-large under NaN, ModernBERT-large
underperforming RoBERTa-large, and the LLM alternative that did not
out-perform RoBERTa-large under cross-validation within our compute
envelope).}
\label{fig:citgraph}
\end{figure}

Figure~\ref{fig:citgraph} traces the prior work that fed each component of
our pipeline. The horizontal axis is left-to-right: citations on the left,
the pipeline stages they motivated on the right. Solid arrows mark routes
that ended up in the submitted system; dashed arrows mark routes we tried
and abandoned (the DeBERTa-v3-large NaN failures, ModernBERT-large
underperformance, and the LLM alternative that did not validate under CV).

\end{document}